\documentclass{mva_style}
\usepackage{graphicx}

\usepackage{url}
\usepackage[hidelinks]{hyperref}

\usepackage{xcolor, soul}

\def \affiliationHHI {Fraunhofer HHI}
\def \affiliationHU {Humboldt University of Berlin}

\finalcopy 

\begin{document}
\title{Automatic Reconstruction of Semantic 3D Models from 2D Floor Plans}

\author{
  Aleixo Cambeiro Barreiro\\
  \affiliationHHI\\
  {\tt aleixo.cambeiro@hhi.fraunhofer.de}\\
  \and
  Mariusz Trzeciakiewicz\\
  \affiliationHHI\\
  {\tt mariusz.trzeciakiewicz@hhi.fraunhofer.de}\\
  \and
  Anna Hilsmann\\
  \affiliationHHI\\
  {\tt anna.hilsmann@hhi.fraunhofer.de}\\
  \and
  Peter Eisert\\
  \affiliationHU\\
  {\tt eisert@informatik.hu-berlin.de}\\
}

\maketitle

\section*{\centering Abstract}
\textit{
  Digitalization of existing buildings and the creation of 3D BIM models for them has become crucial for many tasks. Of particular importance are floor plans, which contain information about building layouts and are vital for processes such as construction, maintenance or refurbishing. However, this data is not always available in digital form, especially for older buildings constructed before CAD tools were widely available, or lacks semantic information. The digitalization of such information usually requires manual work of an expert that must reconstruct the layouts by hand, which is a cumbersome and error-prone process. In this paper, we present a pipeline for reconstruction of vectorized 3D models from scanned 2D plans, aiming at increasing the efficiency of this process. The method presented achieves state-of-the-art results in the public dataset CubiCasa5k~\cite{kalervo2019cubicasa5k}, and shows good generalization to different types of plans. Our vectorization approach is particularly effective, outperforming previous methods.
}

\section{Introduction}

The digitalization of buildings is very important for their construction, maintenance and refurbishing, which due to technological advances depend more and more on digital data. Especially important is the digitalization of floor plans, which allows their use and modification by means of CAD tools or the generation of BIM models (Building Information Modelling), greatly simplifying design and planning processes. However, plans are often not in digital form or do not contain semantic information, especially for older buildings constructed before CAD tools were widely available. In such cases, architects have to resort to manually reproducing the printed plans using a design tool, which is a very monotonous, error-prone and time-consuming effort. This makes the automation of this process a very attractive goal, with potential to save a lot of valuable resources. This is, however, a challenging task given the diversity of symbols, annotation systems, scales, etc. that are present in building floor plans.

Traditionally, handcrafted features and techniques such as morphological image operations, Hough transforms~\cite{hough1962method} or connected components analysis have been used to tackle this problem, as in~\cite{de2014statistical,ahmed2011improved}. However, with recent technological advances brought by artificial neural networks, great improvements have been made in various areas of computer vision, proving them vastly superior to traditional methods in many cases. This has caused a shift towards deep-learning methods in the scientific community, as seen in~\cite{liu2017raster,kalervo2019cubicasa5k,lv2021residential}, which are state-of-the-art papers in the field.

Kalervo et al.~\cite{kalervo2019cubicasa5k} have based their work on the method proposed by Liu et al.~\cite{liu2017raster}, improving upon it and applying it to the CubiCasa5k dataset. They introduce a pipeline in which they first use a CNN to generate segmentation masks from which to extract junctions and corners from walls and symbols, which they later transform into some predefined primitives using integer programming, yielding an output that is postprocessed into the final vectorial representation of the floor layout. They have publicly released their code and the dataset, making direct performance comparisons feasible. For this reason, we have taken the results of this method as a reference in our study.

Lv et al.~\cite{lv2021residential} propose a pipeline in which they use a CNN to extract ROIs containing the area occupied by the floor layout within each image, on which they focus for the next steps. They then use a segmentation network to extract structural elements, including walls, rooms, windows and doors, to which they apply a vectorization algorithm. Additionally, they use a detection model to extract symbols and text, and use this information and the length of measuring lines to calculate the scale of the plans. This method makes some improvements with respect to~\cite{kalervo2019cubicasa5k}, e.g.~the possibility to process inclined walls. However, the code has not been released and some of the data required to train such a method is not available in the public datasets considered, e.g.~annotations of text and measuring lines or annotations of the ROIs of the plans. Our paper draws some ideas from this approach but unfortunately, due to the unavailability of their code and data, cannot be directly compared to it.

In this paper, we propose a pipeline for 3D model reconstruction from 2D floor plans, dividing this process into the following subtasks. As a first stage, we segment the structural elements (walls) and detect relevant symbols such as doors or windows. Then, we extract the wall segments as polygons from the predicted segmentation masks in order to create semantic elements for further usage in CAD tools. Finally, we create a 3D model of the building using the extracted information, which can be exported as IFC model. These steps will be detailed in the following section. Furthermore, we show that our method achieves state-of-the-art results on the reference dataset and that it generalizes well to plans different from those in the dataset.

\section{3D Model Reconstruction Pipeline}

In this section, we discuss our approach for the reconstruction of 2D floor plans, including the different stages of the pipeline to transform them to 3D models. The emphasis will be on the wall segmentation and extraction due to its intrinsic complexity, stemming from factors such as their irregular shapes and sizes, the ambiguity of their subdivision into segments, the discontinuities in their representation to include doors or windows or their similarity to other annotation lines.

\begingroup

\setlength{\tabcolsep}{2pt} 
\def \figWidth {40mm}

\begin{figure*}[!h]
  \begin{tabular}{cccc}
    \includegraphics[width=\figWidth]{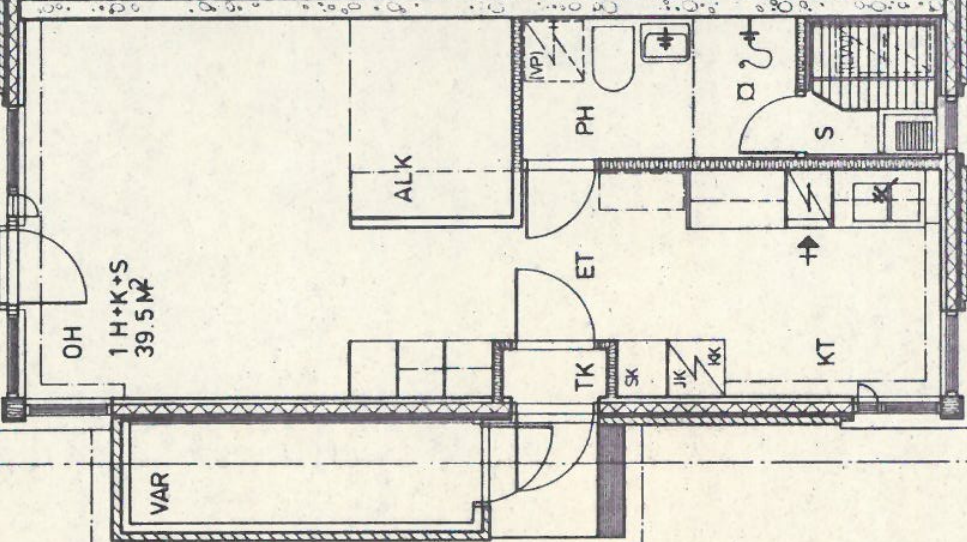} & \includegraphics[width=\figWidth]{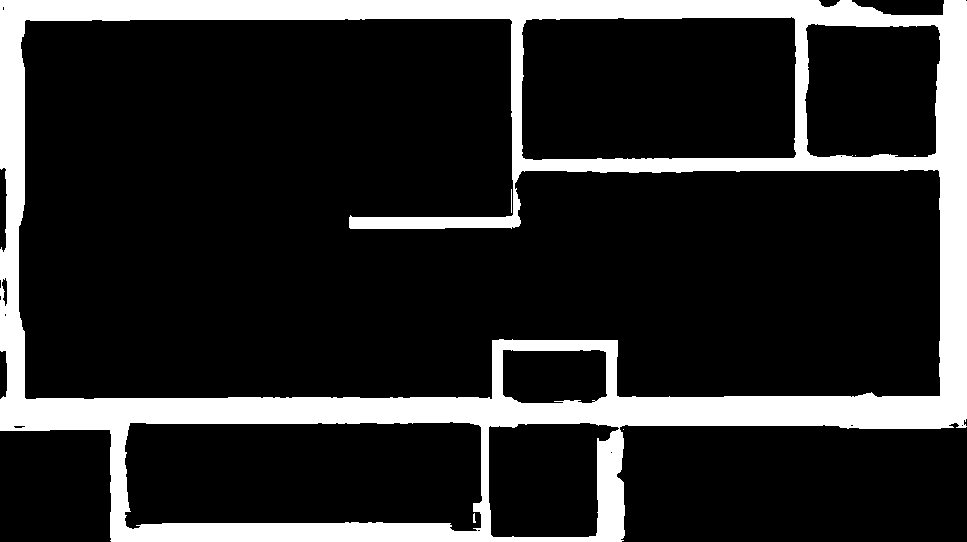} & \includegraphics[width=\figWidth]{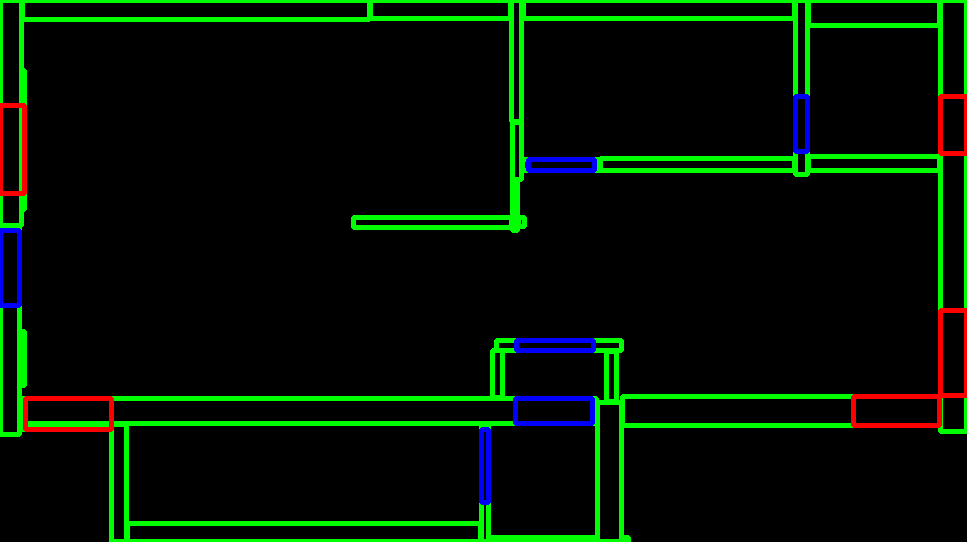} & \includegraphics[width=\figWidth]{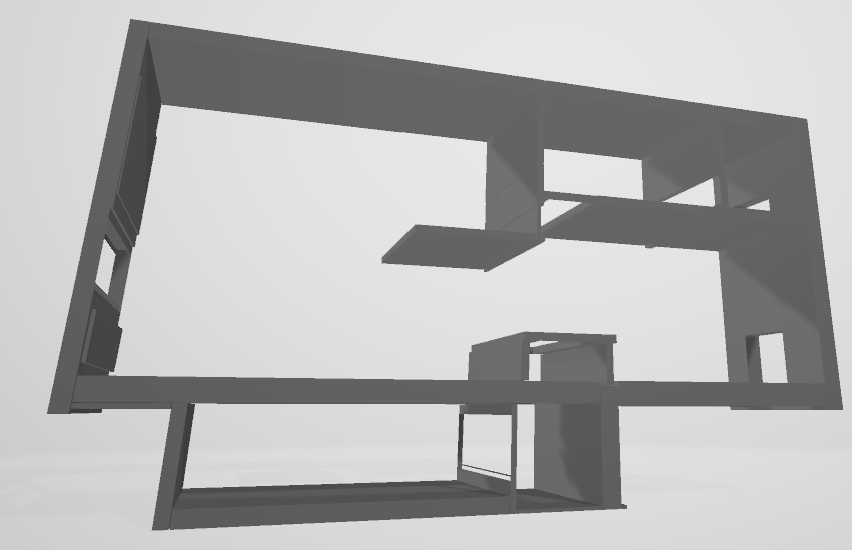} \\
    \includegraphics[width=\figWidth]{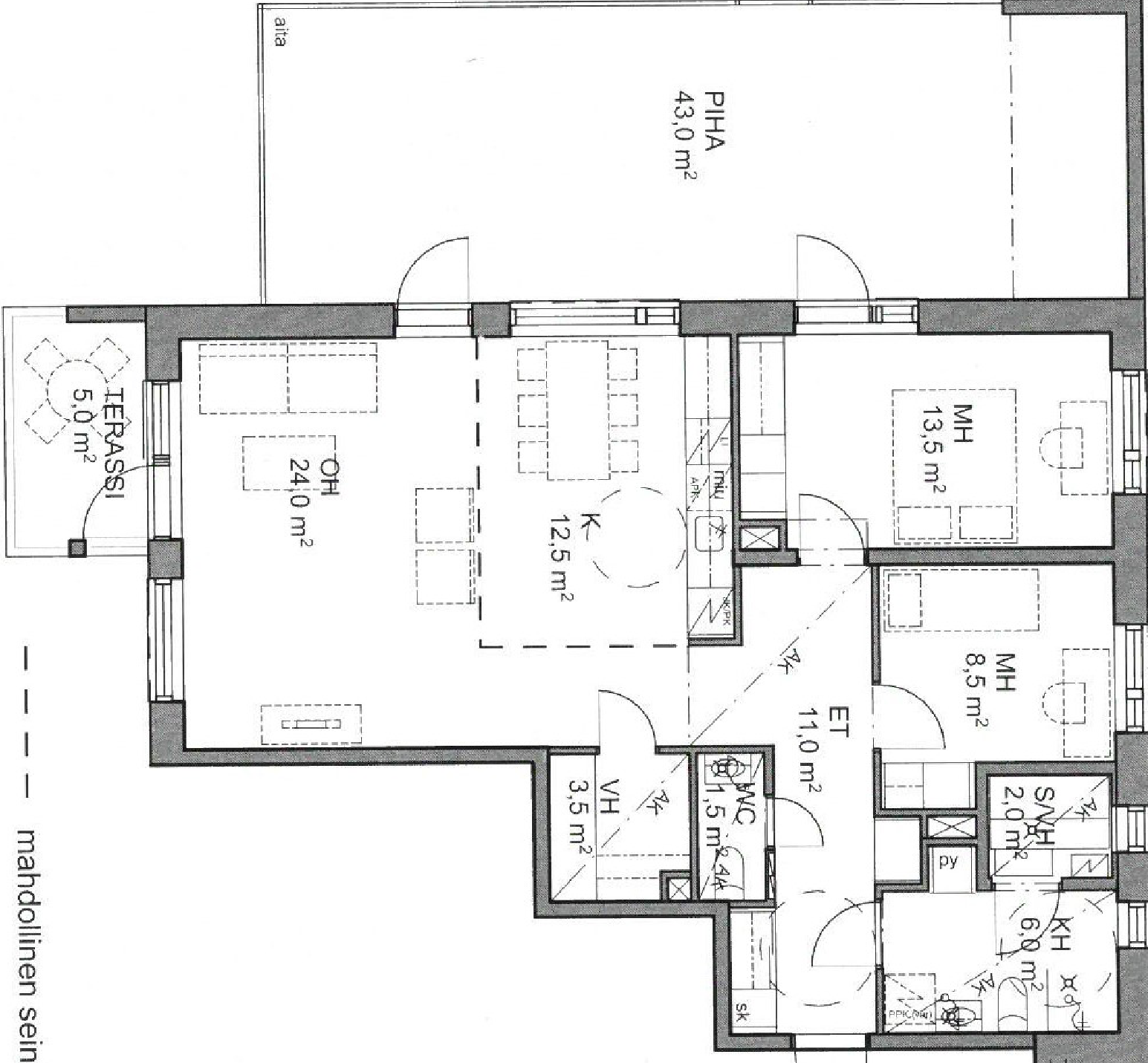} & \includegraphics[width=\figWidth]{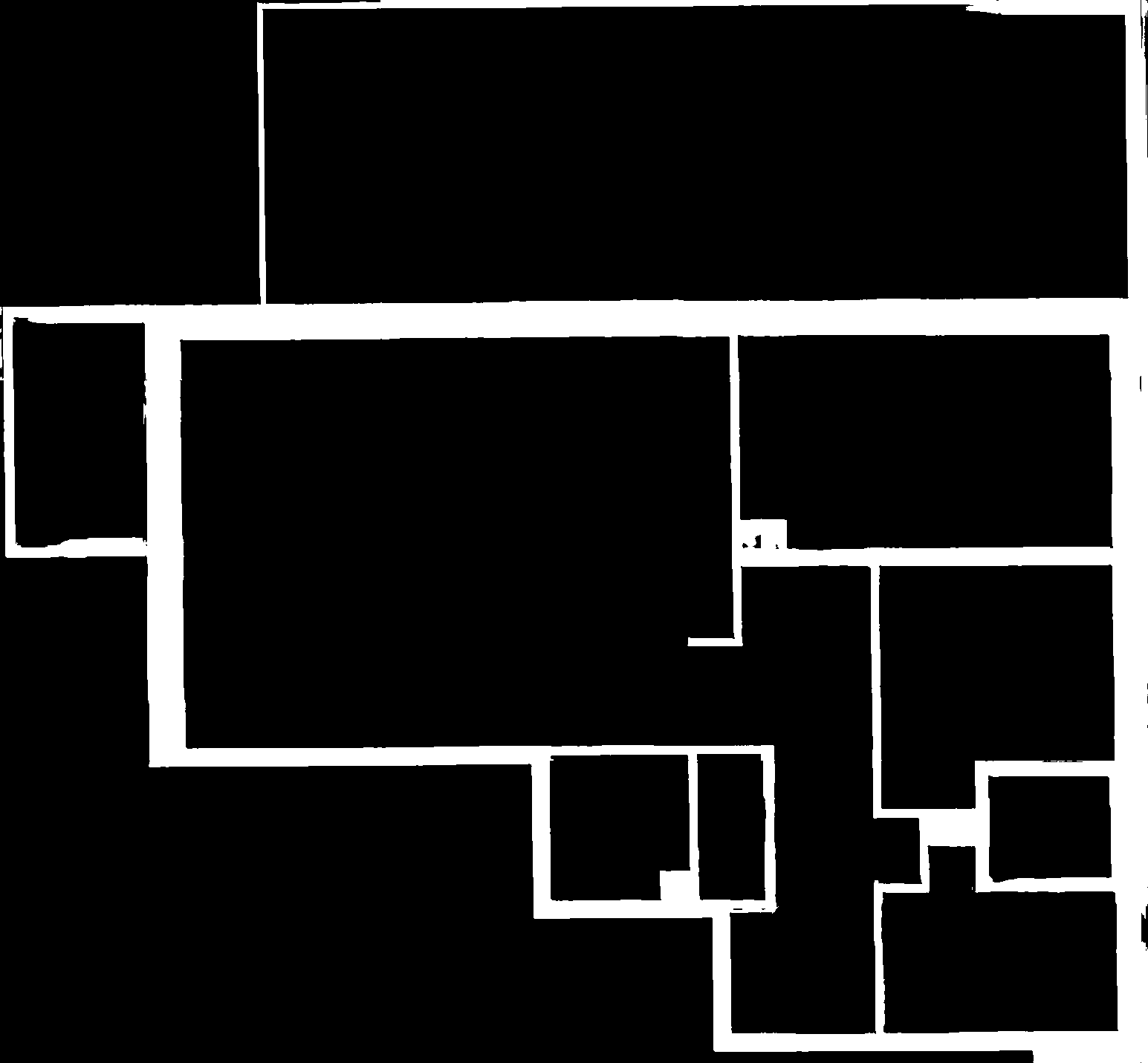} & \includegraphics[width=\figWidth]{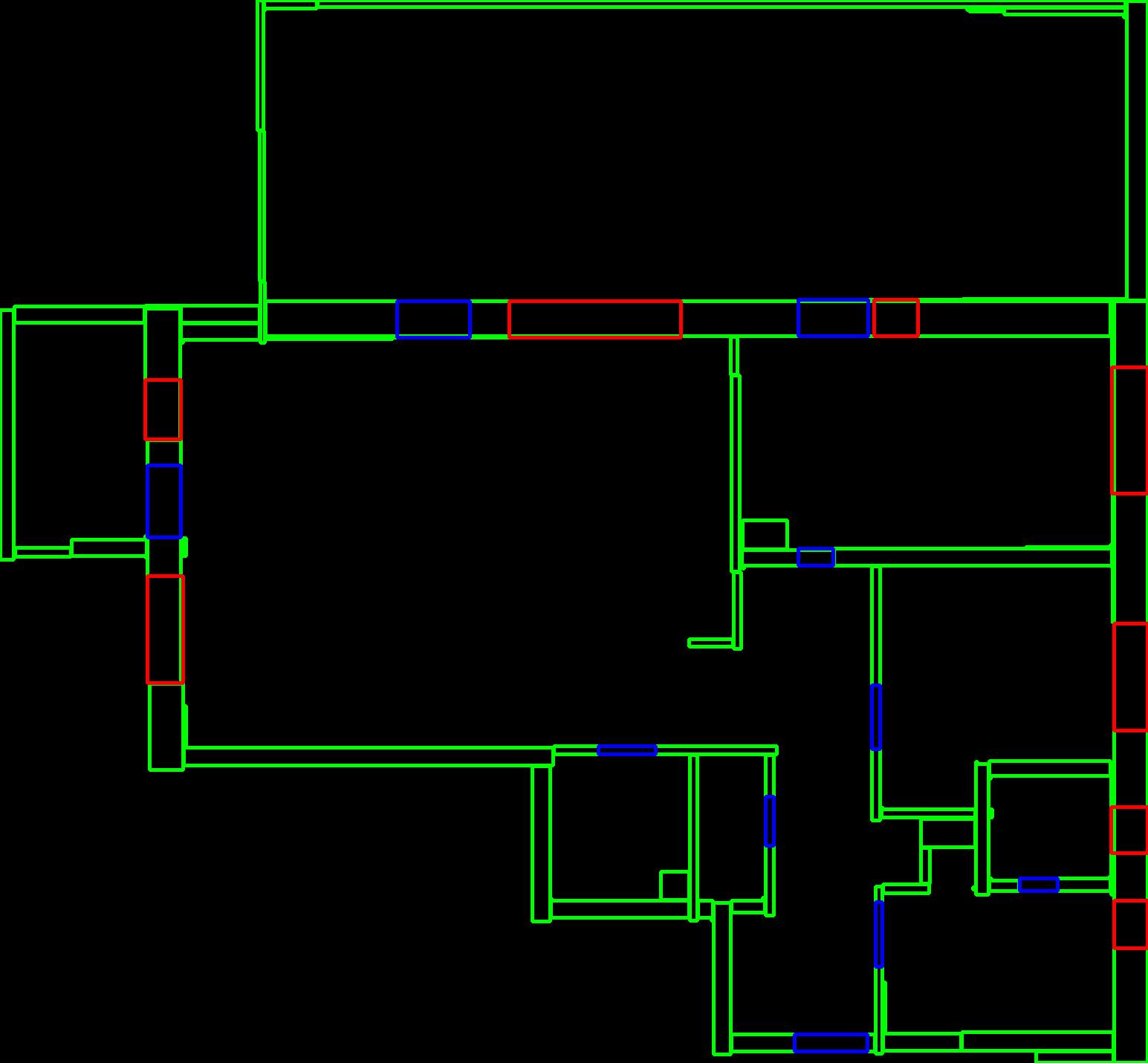} & \includegraphics[width=\figWidth]{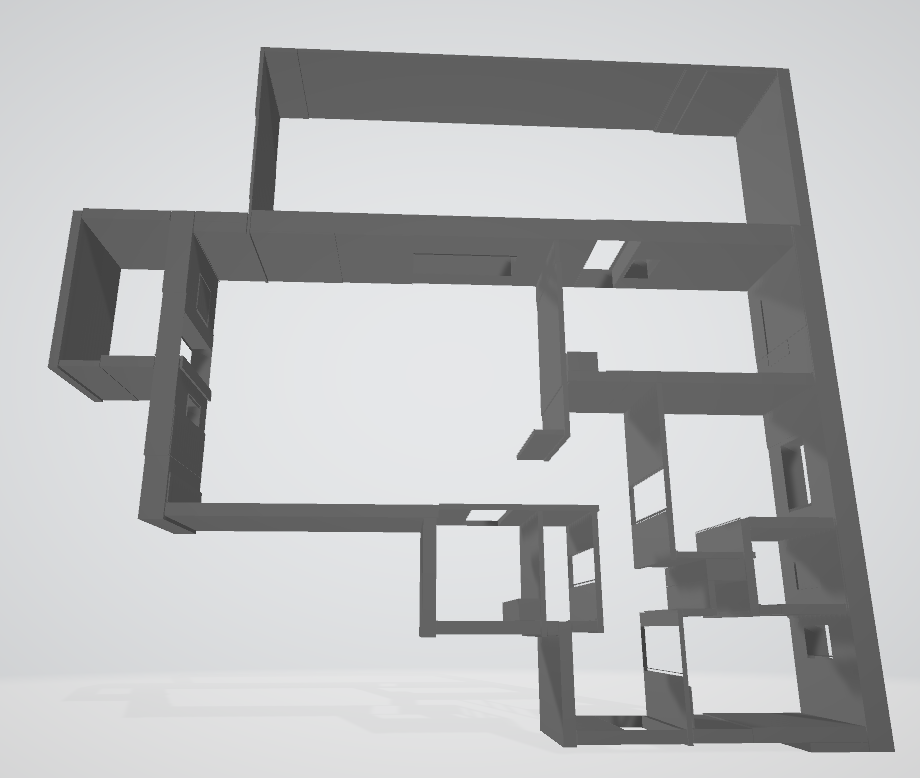} \\
    \includegraphics[width=\figWidth]{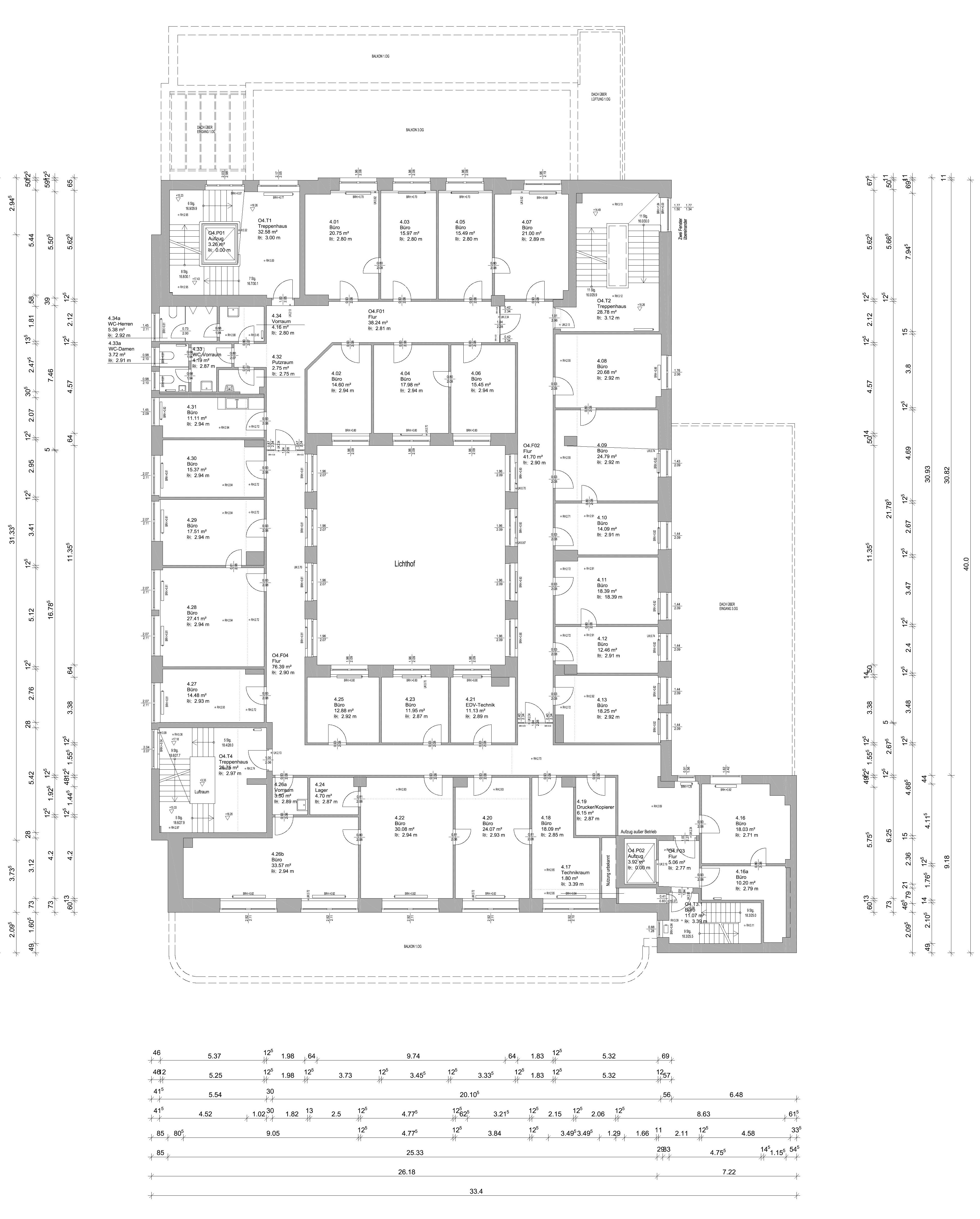} & \includegraphics[width=\figWidth]{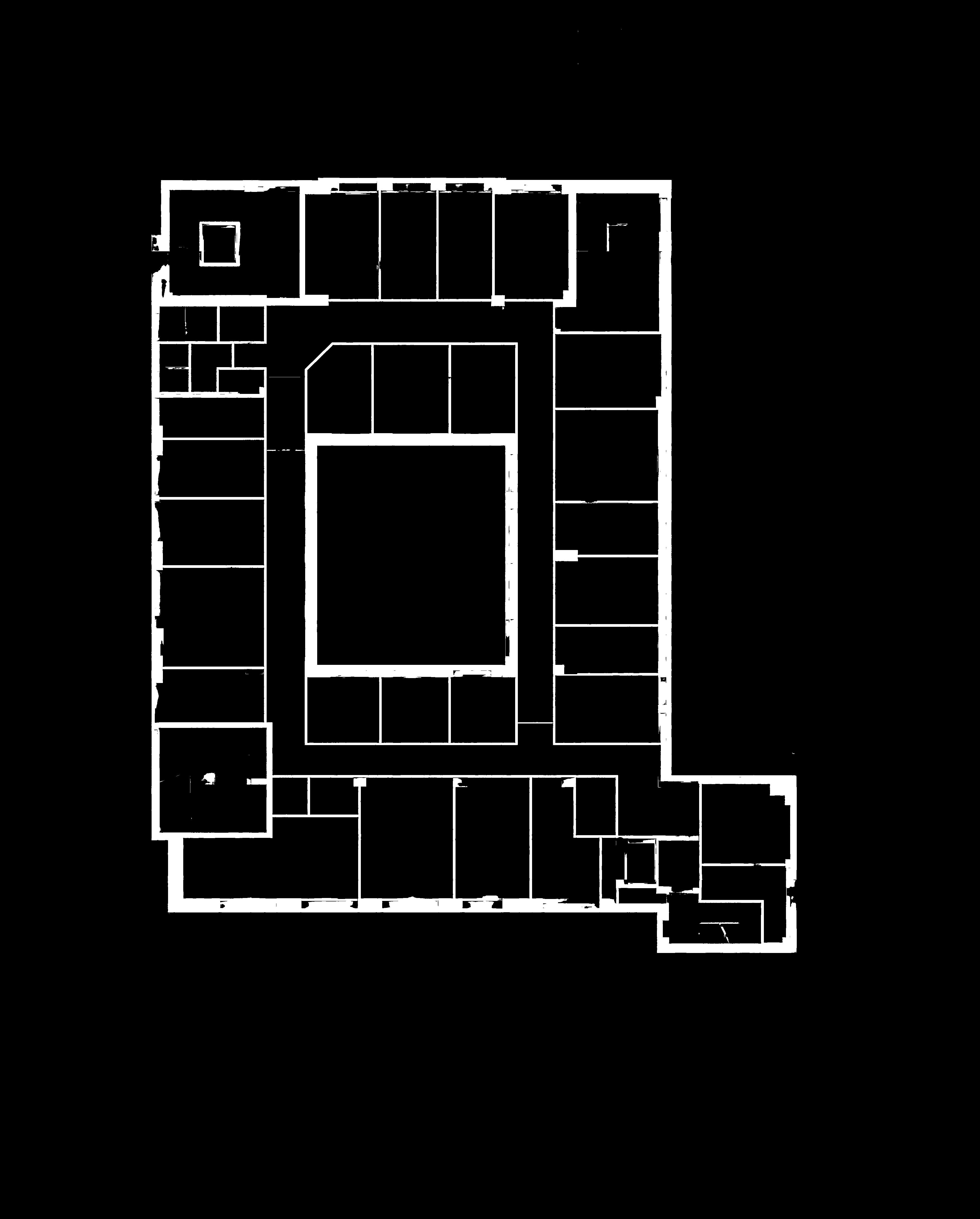} & \includegraphics[width=\figWidth]{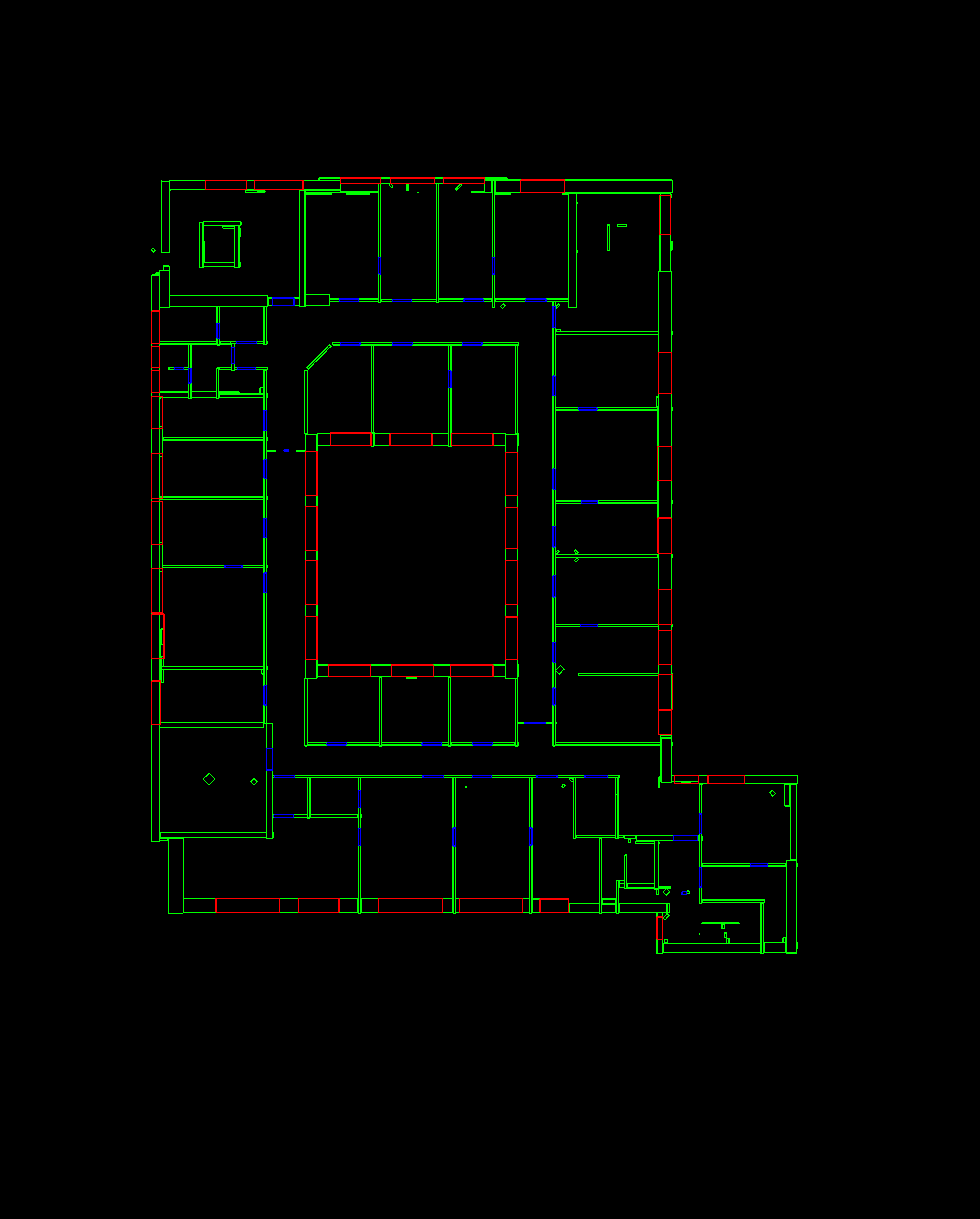} & \includegraphics[width=\figWidth]{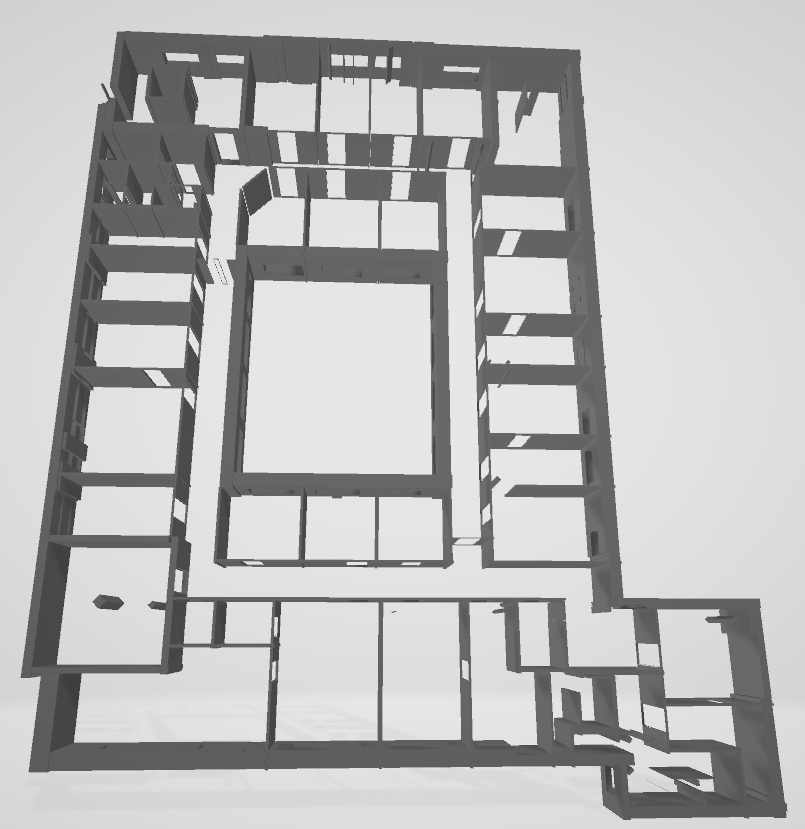} \\
    (a) & (b) & (c) & (d) \\
  \end{tabular}
  \\
  \caption{Different stages of the reconstruction pipeline shown for two images from the validation split of the dataset (top rows) and for an office building plan obtained from a different source (bottom row). Column (a) shows the original plans, (b) shows the predicted segmentation masks, (c) shows the extracted vectorized results (walls in green, doors in blue and windows in red), and (d) shows visualizations of the reconstructed 3D models.}
  \label{fig:pipeline-results}
\end{figure*}

\endgroup

\subsection{Preliminaries}

In our work, we focus mainly on the walls, doors and windows, which are arguably the most important elements in reconstructing a floor layout. We found the rooms annotations to be relatively specific to the type of plans in the dataset, i.e.~residential plans, and therefore perhaps not so useful in generalization to other structures, e.g.~office buildings.

In order to improve the generalization of our method and enable it to deal with images of different sizes, we use a sliding window approach to process the plans with the deep learning models. Additionally, we use diverse data augmentation techniques during training in order to make our system more resilient towards changes in appearance of these plans. These include random scaling, to complement our sliding window approach in dealing with images of different sizes, and random rotations to help overcome the trend of mostly axis-aligned, vertical and horizontal structures present in the dataset.

\subsection{Symbol detection}

For the symbol detection task, we have opted for a bounding-box detection neural network. This seems to be a suitable tool, since the construction element symbols are discrete, repeated patterns for which an encasing rectangular box can be easily defined. For the architecture, we have chosen the popular Faster-RCNN~\cite{ren2015faster} model with a ResNet~\cite{he2016deep} backbone, which is widely used for object detection tasks due to its good performance. In this case, the focus is on doors and windows, since they are the most relevant in defining the floor layout of a building, allowing for its 3D reconstruction.

\subsection{Wall segmentation}

For the wall detection task, we have chosen a semantic-segmentation approach due to the variability and irregularity in shape and size. We use an FPN~\cite{lin2017feature} architecture with a ResNet~\cite{he2016deep} backbone. Similarly to Lv et al.~\cite{lv2021residential}, for training we have chosen to add to the binary cross-entropy loss an affinity-field loss~\cite{ke2018adaptive}, which is a neighborhood-based loss function aiming at improving the definition of object borders. To avoid fine-tuning the weights of each loss, we follow the method of Kendall et al.~\cite{kendall2018multi}, and learn these weights simultaneously during training.

\subsection{Wall extraction}

Our wall segmentation approach produces pixelwise binary masks with visible outlines of walls and rooms. However, in order for these masks to be used by CAD software they must be vectorized. To this end, we extract individual wall segments and find suitable rectangular box representations for them. The process we follow is described in the following steps.

Firstly, as a preprocessing step, some noise is removed with a morphological opening filter, the image is smoothed with Gaussian blur, and small holes within walls are filled with the help of a morphological closing filter. 

Afterwards, we find angles at which the walls within a mask are rotated. Most walls in reference dataset~\cite{kalervo2019cubicasa5k} are either horizontally or vertically oriented. However, it is also important to consider inclined walls for generalization purposes. In order to find these angles, we extract edges using a Canny operator~\cite{canny1986computational} and apply a Hough transform~\cite{hough1962method} to the resulting image, allowing us to calculate a histogram of line orientations. 

For each angle, we rotate the mask to make wall segments in this direction horizontal or vertical and compute two images, one containing only vertical segments of the rotated mask and one with only horizontal segments. To this end, we use a morphological opening filter with horizontal and vertical kernels. This means that 4 angles can be processed simultaneously, e.g.~by processing 0\textdegree, we already consider angles 90\textdegree, 180\textdegree, 270\textdegree, and 360\textdegree. Then, we apply a contour extraction algorithm~\cite{SUZUKI198532}  to obtain individual wall segments. However, the morphological opening operation that we use to obtain horizontal and vertical components sometimes struggles with slightly inclined walls and considers some of them horizontal/vertical. Therefore, we implement a validation algorithm that removes tilted walls. The algorithm calculates a minimum-area rotated bounding box of a wall and uses this angle to determine whether the wall is tilted and should be removed. Finally, we subtract found walls from the original mask, so that only inclined walls remain. These walls can be then extracted in further iterations with their corresponding angles.

The most important step is to transform irregularly shaped walls into rectangles with a minimal loss of information about the wall's original outline. To this purpose, we have implemented a shrinking algorithm. The algorithm first calculates an axis-aligned bounding box of a wall. Then, it creates four new boxes, each shrunk by one pixel from a different direction (top, left, down, right). Note that new boxes are not rotated, which is why we cannot process inclined walls in this step. For each new box, we calculate intersection over union between the box and the original wall. The box with the biggest intersection over union value becomes the new bounding box for the wall if intersection over union has improved. We repeat these steps until we reach the desired intersection over union threshold or the box becomes too small. If some bigger parts of the original wall lie outside of the bounding box borders, we apply the shrinkage process again to each remaining chunk separately.

Through the vectorization process, resulting boxes can intersect with each other. Therefore, we iterate over extracted boxes to find overlapping ones. If a box is completely surrounded by a bigger box, the smaller one is removed. If the boxes are just partially overlapping, we shrink one of the boxes to remove the intersection. To this end, we chose the box that will result in a smaller overall information loss when shrunk.

\subsection{3D Reconstruction}

Our goal is to build a 3D model from the extracted information. For this purpose, we assume the heights of walls, doors and windows. For future work, this information could be included from other sources, e.g. textual information present in the plans could be detected and analyzed for height references.
Using these assumptions and the extracted information, we generate a 3D model based on rectangular box primitives for the walls, clipping out holes for the doors and windows. In order to do this, the detected doors and windows are matched to their corresponding wall segments, adapting their depths and orientations accordingly when needed. Due to the previous vectorization of the results, the wall segment with the highest overlap with the door or window can easily be determined and the extension of the latter can be adapted to fit the corresponding wall segment.

\section{Experimental results}

Performance of artificial neural networks is heavily dependent on the amount and quality of annotated training data available. Due to data privacy concerns, floor plans of buildings are usually not public domain and annotated datasets are scarce and difficult to obtain. Some publicly available datasets are CVC-FP~\cite{de2015cvc}, Rent3D\cite{liu2015rent3d} and the annotated subset of the LIFULL HOME dataset~\cite{lifull} proposed by Liu et al.~\cite{liu2017raster}. However, these datasets are not ideal for our task due to their insufficient size and other limitations. Lv et al.~\cite{lv2021residential} create a more complete dataset by means of web crawling and manual annotation, which has, however, not been publicly released. CubiCasa5K~\cite{kalervo2019cubicasa5k} is, to the best of our knowledge, the most suitable public dataset for this task, including 5000 annotated plans, divided into a train and a validation split. Hence, for our experiments we used this dataset for training and evaluation. This is a challenging dataset due to the variability in appearance of its images, including many scanned plans with manual modifications or even some drawn by hand entirely. We have noticed that some of the images include annotations for only some part of the visible elements of the plan. We address this by cropping the images and the corresponding labels to the size of the wall masks on the ground truth, leaving out in most cases the elements that have not been annotated and improving the overall quality of the dataset as training material.

\begin{table}[t]
  \caption{Evaluation results on the CubiCasa5k dataset. IoU scores are shown for the raw segmentation masks and the vectorized output of each approach. B represents the baseline, C, training on the cropped dataset, OV, using our vectorization approach and Ours is our method (including C + OV).}
  \begin{center}
    \begin{tabular}{c | c c}
      \hline
      \hline
      \makebox[10mm]{} & \makebox[15mm]{IoU mask} & 
      \makebox[15mm]{IoU vect.}\\
      \hline
      B \cite{kalervo2019cubicasa5k} & 0.74 & 0.53 \\
      B + C & 0.79 & 0.58 \\
      B + C + OV & 0.79 & 0.78 \\
      Ours & \textbf{0.81}   & \textbf{0.8}\\
      \hline
      \hline
    \end{tabular}
    \label{tab:results_cubi_vs_our}
  \end{center}
\end{table}

For the quantitative evaluation of our experiments, we mainly focused on the quality of the produced structural layout, i.e. the extracted walls, since it is most relevant to our problem and the most challenging task in achieving a good reconstruction of the plan. We compare the results on the reference dataset of the method proposed in~\cite{kalervo2019cubicasa5k}, from here on referred to as \textit{baseline} (B), with those of our wall segmentation and extraction model. To achieve a quantitative comparison of both methods, \emph{mean intersection over union} (IoU) between the ground truth and the predicted mask is chosen as a metric. The IoU is also compared for the results of the vectorization methods, which represent our final output. We trained the baseline model twice, both using the dataset in its unaltered form and using our cropped version of it (C), described previously in this section. Furthermore, we have applied our own wall vectorization algorithm (OV) to raw segmentation masks produced by the baseline approach. Table~\ref{tab:results_cubi_vs_our} presents the results of the experiment, showing a clear improvement in the vectorization process of our method over the baseline.

For a qualitative analysis of the results, Figure~\ref{fig:pipeline-results} shows different stages of the pipeline for some sample images from the validation set. Additionally, our method has been tested with some plans we acquired that are not part of the training set. These are office buildings plans with sizes, styles, overall layouts and symbols largely different to the ones present on CubiCasa5k. Such sample results are also shown in Figure~\ref{fig:pipeline-results}.

\section{Conclusion}

In this paper, we present a pipeline for reconstruction of 3D models from 2D floor plans of buildings. Our method achieves good results on the CubiCasa5k~\cite{kalervo2019cubicasa5k} dataset, outperforming prior methods like \cite{kalervo2019cubicasa5k}. Especially our new vectorization increases quality with respect to previous methods. Furthermore, we show that our method generalizes well to other floor plans with largely different characteristics, such as plans for office buildings with styles and sizes completely unlike the plans seen in the dataset, obtaining good reconstruction results for the plans tested.

\section*{Acknowledgements}
This work is supported by the German Federal Ministry for Economic Affairs and Climate Action (BIMKIT, grant no. 01MK21001H).

\bibliographystyle{abbrv}
\bibliography{bibliography}






\end{document}